# Evidence as Opinions of Experts

by

*Robert A. Hummel* †
Courant Institute of Mathematical Sciences

*Michael S. Landy* ‡
Department of Psychology

†New York University
251 Mercer Street
New York, New York 10012

‡New York University
6 Washington Place
New York, New York 10003

Presented at the "Uncertainty in AI" Workshop,
Philadelphia, PA, August 8 - 10, 1986

This research was supported by Office of Naval Research Grant N00014-85-K-007 and NSF Grant DCR-8403300. This is a conference report version of a more complete paper, which appears as NYU Robotics Research Report #57, submitted to IEEE-Transactions on Pattern Analysis and Machine Intelligence.

135

# Evidence as Opinions of Experts

*Robert Hummel*
*Michael Landy*

Courant Institute of Mathematical Sciences
New York University

## Abstract

We describe a viewpoint on the Dempster/Shafer "Theory of Evidence", and provide an interpretation which regards the combination formulas as statistics of the opinions of "experts". This is done by introducing spaces with binary operations that are simpler to interpret or simpler to implement than the standard combination formula, and showing that these spaces can be mapped homomorphically onto the Dempster/Shafer theory of evidence space. The experts in the space of "opinions of experts" combine information in a Bayesian fashion. We present alternative spaces for the combination of evidence suggested by this viewpoint.

## 1. Introduction

Many problems in artificial intelligence call for assessments of degrees of belief in propositions based on evidence gathered from disparate sources. It is often claimed that probabilistic analysis of propositions is at variance with intuitive notions of belief [1,2,3]. Various methods have been introduced to reconcile the discrepancies, but no single technique has settled the issue on both theoretical and pragmatic grounds.

One method for attempting to modify probabilistic analysis of propositions is the Dempster/Shafer "Theory of Evidence." This theory is derived from notions of upper and lower probabilities, as developed by Dempster in [4]. The idea that intervals instead of probability values can be used to model degrees of belief had been suggested and investigated by earlier researchers [5,6,2,7], but Dempster's work defines the upper and lower points of the intervals in terms of statistics on set-valued functions defined over a measure space. The result is a collection of intervals defined for subsets of a fixed labeling set, and a combination formula for combining collections of intervals.

Dempster explained in greater detail how these notions could be used to assess beliefs on propositions in [8]. The topic was taken up by Shafer [9,10], and led to publication of a monograph on the "Theory of Evidence," [11]. All of these works after [8] emphasize the values assigned to subsets of propositions (the "beliefs"), and the combination formulas, and de-emphasize the connection to the statistical foundations based on the set-valued functions on a measure space. This paper will return to the original formulation by Dempster in [8] to relate the statistical foundations of the Dempster/Shafer theory of evidence to notions of beliefs on propositions.

This paper has three main points. First, we show that the combination rule for the Dempster/Shafer theory of evidence may be simplified by omiting the normalization term. We next point out that the individual pairs of experts involved in the combination formula can be regarded as performing Bayesian updating. Finally, we present extensions to the theory, based on allowing experts to express probabilistic opinions and assuming that the logarithms of experts' opinions over the set of labels are multi-normally distributed.

## 2. The Rule of Combination and Normalization

The purpose of this section is to show how one can dispense with the normalization term in the Dempster rule of combination.

The set of possible outcomes, or labelings, will be denoted in this paper by $\Lambda$. This set is the "frame of discernment", and in other works has been denoted, variously, by $\Omega$, $\Theta$, or $S$. For convenience, we will assume that $\Lambda$ is a finite set with $n$ elements, although the framework could easily be extended to continuous label sets. More importantly, we will assume that $\Lambda$ represents a set of states that are mutually exclusive and exhaustive.

An element (or state of belief) in the theory of evidence is represented by a probability distribution over the power set of $\Lambda$, $P(\Lambda)$. That

is, a state $m$ is

$$m : P(\Lambda) \to [0,1], \quad \sum_{A \subseteq \Lambda} m(A) = 1. \quad (1a)$$

There is an additional proviso that is typically applied, namely that every state $m$ satisfies

$$m(\emptyset) = 0. \quad (1b)$$

Section 3.2 introduces a plausible interpretation for the quantities comprising a state.

A state is updated by combination with new evidence, or information, which is presented in the form of another state. Thus given a current state $m_1$, and another state $m_2$, a combination of the two states is defined to yield a state $m_1 \oplus m_2$ which for $A \neq \emptyset$ is given by

$$(m_1 \oplus m_2)(A) = \frac{\sum_{B \cap C = A} m_1(B) m_2(C)}{1 - \sum_{B \cap C = \emptyset} m_1(B) m_2(C)} \quad (2a)$$

and is zero for $A = \emptyset$. This is the so called "Dempster Rule of Combination."

The problem with this definition is that the denominator in (2a) might be zero, so that $(m_1 \oplus m_2)(A)$ is undefined. That is, there exist pairs $m_1$ and $m_2$ such that the combination of $m_1$ and $m_2$ is not defined. This, of course, is not a very satisfactory situation for a binary operation on a space. The solution which is frequently taken is to avoid combining such elements. An alternative is to add an additional element $m_0$ to the space:

$$m_0(A) = 0 \text{ for } A \neq \emptyset, \quad m_0(\emptyset) = 1.$$

Note that this additional element does not satisfy the condition $m(\emptyset) = 0$. Then define, as a special case,

$$m_1 \oplus m_2 = m_0 \text{ if } \sum_{B \cap C = \emptyset} m_1(B) m_2(C) = 1. \quad (2b)$$

The binary operation is then defined for all pairs $m_1, m_2$. The special element $m_0$ is an absorbent state, in the sense that $m_0 \oplus m = m \oplus m_0 = m_0$ for all states $m$.

**Definition 1:** We define $(M, \oplus)$, the *space of belief states*, by $M = \{m \text{ satisfying (1a) and (1b)}\} \cup \{m_0\}$, and define $\oplus$ by (2a) when the denominator in (2a) is nonzero, and by (2b) otherwise. ∎

The set $M$, together with the combination operation $\oplus$, constitutes a *monoid*, since the binary operation is closed and associative, and there is an identity element. In fact, the binary operation is commutative, so we can say that the space is an abelian monoid.

Still, because of the normalization and the special case in the definition of $\oplus$, the monoid $M$ is both ugly and cumbersome. It makes better sense to dispense with the normalization. We have

**Definition 2:** We define $(M', \oplus')$, the *space of unnormalized belief states*, by $M' = \{ m \text{ satisfying (1a)} \}$ without the additional proviso, and set

$$(m_1 \oplus' m_2)(A) = \sum_{B \cap C = A} m_1(B) \cdot m_2(C) \quad (3)$$

for all $A \subseteq \Lambda$ and for all pairs $m_1, m_2 \in M'$. ∎

One can verify that $m_1 \oplus' m_2 \in M'$, and that $\oplus'$ is associative and commutative, and that there is an identity element. Thus $M'$ is also an abelian monoid. Clearly, $M'$ is a more attractive monoid than $M$.

We define a transformation $V$ mapping $M'$ to $M$ by the formulas

$$(Vm)(A) = \frac{m(A)}{1 - m(\emptyset)}, \quad (Vm)(\emptyset) = 0 \quad (4)$$

if $m(\emptyset) \neq 1$, and $Vm = m_0$ otherwise.

A computation shows that $V$ preserves the binary operation; i.e.,

$$V(m_1 \oplus' m_2) = V(m_1) \oplus V(m_2).$$

Thus $V$ is a *homomorphism*. Further, $V$ is onto, since for $m \in M$, the same $m$ is in $M'$, and $Vm = m$. The algebraic terminology is that $V$ is an *epimorphism* of monoids, a fact which we record in

**Lemma 1:** $V$ maps homomorphically from $(M', \oplus')$ onto $(M, \oplus)$. ∎

A "representation" is a term that refers to a map that is an epimorphism of structures. Intuitively, such a map is important because it allows us to consider combination in the space formed by the range of the map as combinations of preimage elements. Lemma 1 will eventually form a small part of a representation to be defined in the next section.

In the case in point, we see that combination can be done without a normalization factor. If it is required to combine elements in $M$, one can perform the combinations in $M'$, and project to $M$ by $V$ after all of the combinations are completed. In terms of the Dempster/Shafer theory of evidence, this result says that the normalization in the combination formula is essentially irrelevant, and that combining can be handled by Equation (3) in place of Equation (2a).

### 3. Spaces of Opinions of Experts

In this section, we introduce two new spaces, based on the opinions of sample spaces of experts, and discuss the evaluation of statistics of experts opinions. Finally, we interpret the combination rules in these spaces as being a form of Bayesian updating. In the following section we will show that these spaces also map homomorphically onto the space of belief states. Thus our intent is to show that the Dempster/Shafer space of belief states can be interpreted as the statistics of experts updating their opinions in a Bayesian fashion. The reason why the formulas don't look like Bayesian updating is that instead of having a single expert, there are collections of experts, updating in pairs. Thus instead of keeping track of the opinion of a single expert receiving evidence from different sources, we will see that the space of beliefs can be viewed as the statistics of collections of experts, combining their opinions in a Bayesian fashion, where each collection of experts represents an independent source of information. We begin by giving a formal introduction to the spaces of expert and their methods of combination.

### 3.1. Opinions of Experts

We consider a set $\mathcal{E}$ of "experts", together with a map $\mu$ giving a weight or strength for each expert. It is convenient to think of $\mathcal{E}$ as a large but finite set, although the essential restriction is that $\mathcal{E}$ should be a measure space. Each expert $\omega \in \mathcal{E}$ maintains a list of possible labels: Dempster uses the notation $\Gamma(\omega)$ for this subset; i.e., $\Gamma(\omega) \subseteq \Lambda$. Here we will assume that each expert $\omega$ has more than just a subset of possibilities $\Gamma(\omega)$, but also a *probabilistic opinion* $p_\omega$ defined on $\Lambda$, such that $p_\omega(\lambda)$ is a probability distribution over $\lambda \in \Lambda$. The value $p_\omega(\lambda)$ represents expert $\omega$'s assessment of the probability of occurrence of the label $\lambda$. Except in the case that $\omega$ has no opinion (see below), we necessarily have

$$\sum_{\lambda \in \Lambda} p_\omega(\lambda) = 1.$$

If an expert $\omega$ believes that a label $\lambda$ is possible, i.e., $\lambda \in \Gamma(\omega)$, then the associated probability estimate $p_\omega(\lambda)$ will be nonzero. Conversely, if $\omega$ thinks that $\lambda$ is impossible ($\lambda \notin \Gamma(\omega)$), then $p_\omega(\lambda) = 0$. We also include the possibility that expert $\omega$ has no opinion which is indicated by the special element $p_\omega \equiv 0$. This state is included in order to ensure that the binary operation, to be defined later, is closed. We denote the collection of probabilistic opinions $\{ p_\omega \mid \omega \in \mathcal{E} \}$ by $P$.

It will turn out that the central point in the theory of evidence is that the $p_\omega(\lambda)$ data is used only in terms of test for zero. Specifically, we set

$$x_\omega(\lambda) = \begin{cases} 1 & \text{if } p_\omega(\lambda) > 0 \\ 0 & \text{if } p_\omega(\lambda) = 0. \end{cases} \quad (5)$$

Note that $x_\omega$ is the characteristic function of the set $\Gamma(\omega)$ over $\Lambda$, i.e., $x_\omega(\lambda) = 1$ iff $\lambda \in \Gamma(\omega)$. The collection of all $x_\omega$'s will be denoted by $X$, and will be called the *boolean opinions* of the experts $\mathcal{E}$.

If we regard the space of experts $\mathcal{E}$ as a sample space, then each $x_\omega(\lambda)$ can be regarded as a sample of a random (boolean) variable $x(\lambda)$. In a similar way, the $p_\omega(\lambda)$'s are also samples of random variables $p(\lambda)$. In the next section, we will define the state of the system will be defined by statistics on the set of random variables $\{x(\lambda)\}_{\lambda \in \Lambda}$. These statistics are measured over the space of experts. If all experts have the same opinion, then the state should describe that set of possibilities, and the fact that there is a unanimity of opinion. If there is a divergence of opinions, the state should record the fact. The essential idea here is that we measure uncertainty in a probabilistic or boolean opinion by sampling a variety of opinions among a collection of experts, and observe the spread in those opinions.

A key aspect to the spaces of opinions of experts is that collections of experts are combined by taking product sets of experts. That is, suppose $\mathcal{E}_1$ is one set of experts with their opinions, and $\mathcal{E}_2$ is another set of experts with their opinions. The combination element will have as its set of experts the product set $\mathcal{E}_1 \times \mathcal{E}_2$. It might seem more desirable to make use of the disjoint union of $\mathcal{E}_1$ and $\mathcal{E}_2$, but then the connection with the Dempster/Shafer combination formula would not hold. The statistics of the combination element will depend on the statistics of the constituent elements because combination is defined by taking a product set of experts.

Pairs of experts combine their opinions in an essentially Bayesian fashion. Under fairly standard independence assumptions, two experts should update their probabilistic assignment for a given label by taking the product of their individual probabilities, and dividing by a prior probability. The resulting values have to be normalized so that they remain probabilities. In terms of the boolean opinions, Bayesian updating with the same independence assumption asserts that two experts agree that a label is possible only if both experts believe the label to be possible.

We are now ready to introduce the spaces which we will term "opinions of experts." The central point is that the set of labels $\Lambda$ is fixed, but that the set of experts $\mathcal{E}$ can be different for distinct elements in these spaces.

**Definition 3:** The *space of boolean opinions of experts*, $(\mathcal{N}', \odot)$, is defined by:

$$\mathcal{N}' = \{(\mathcal{E}, \mu, X) \mid \#\mathcal{E} < \infty, \mu \text{ is a measure on } \mathcal{E},$$

$$X = \{x_\omega\}_{\omega \in \mathcal{E}}, \quad x_\omega : \Lambda \to \{0,1\} \; \forall \omega\}.$$

If $(\mathcal{E}_1, \mu_1, X_1)$ and $(\mathcal{E}_2, \mu_2, X_2)$ are elements in $\mathcal{N}'$, define their product

$$(\mathcal{E}, \mu, X) = (\mathcal{E}_1, \mu_1, X_1) \odot (\mathcal{E}_2, \mu_2, X_2)$$

by

$$\mathcal{E} = \mathcal{E}_1 \times \mathcal{E}_2 = \{(\omega_1, \omega_2) \mid \omega_1 \in \mathcal{E}_1, \omega_2 \in \mathcal{E}_2\}$$

$$\mu(\{(\omega_1, \omega_2)\}) = \mu_1(\{\omega_1\}) \cdot \mu_2(\{\omega_2\}),$$

and

$$X = \{x_{(\omega_1,\omega_2)}\}_{(\omega_1,\omega_2) \in \mathcal{E}},$$

$$x_{(\omega_1,\omega_2)}(\lambda) = x_{\omega_1}^{(1)}(\lambda) \cdot x_{\omega_2}^{(2)}(\lambda),$$

where $X_i = \{x_{\omega_i}^{(i)} \mid \omega_i \in \mathcal{E}_i\}$, for $i = 1, 2$. ∎

**Definition 4:** Let $K = \{\kappa_\lambda\}$ be a set of positive constants indexed over the label set $\Lambda$. The *space of probabilistic opinions of experts* $(\mathcal{N}, K, \otimes)$ is defined similarly, except that $\mathcal{N}$ consists of triples $(\mathcal{E}, \mu, P)$, where $P$ contains probabilistic opinions (as introduced earlier), and that combination of those probabilistic opinions is defined by

$$p_{(\omega_1,\omega_2)}(\lambda) = \frac{p_{\omega_1}^{(1)}(\lambda) p_{\omega_2}^{(2)}(\lambda) [\kappa_\lambda]^{-1}}{\sum_{\lambda'} p_{\omega_1}^{(1)}(\lambda') p_{\omega_2}^{(2)}(\lambda') [\kappa_{\lambda'}]^{-1}},$$

providing the denominator is nonzero, and

$$p_{(\omega_1,\omega_2)} \equiv 0$$

otherwise. Here, $P_i = \{p_{\omega_i}^{(i)}\}_{\omega_i \in \mathcal{E}_i}$ for $i = 1, 2$, and the $\kappa_\lambda$'s are a fixed set of positive constants defined for $\lambda \in \Lambda$. ∎

To interpret this combining operation, consider two sets of experts $\mathcal{E}_1$ and $\mathcal{E}_2$, with each set of experts expressing opinions in the form of $P_1$ and $P_2$. We form a new set of experts, which is simply the set of all committees of two, consisting of one expert from $\mathcal{E}_1$, and another from $\mathcal{E}_2$. In each of the committees, the members confer to determine a consensus opinion. In the probabilistic case, one can interpret the formulas as Bayesian combination (where $\kappa_\lambda$ is the prior probability Prob ( $\mathbb{l}$ ) on $\lambda$). In the boolean case, the consensus is simply the intersection of the composing opinions (see Figure 1).

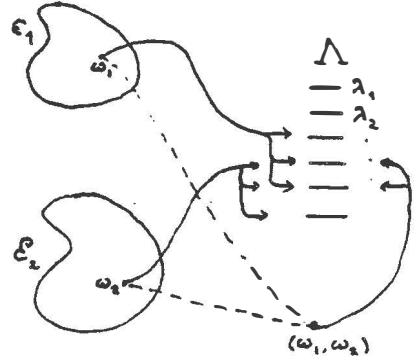

**Figure 1.** A depiction of the combination of two boolean opinions of two experts, yielding a consensus opinion by the element in the product set of experts formed by the committee of two.

### 3.2. Statistics of Experts

It will turn out that the Dempster/Shafer theory of evidence can be viewed as a mechanism for tracking statistics in the space of opinions of experts. We accordingly now define what is meant by the statistics of an element $(\mathcal{E}, \mu, X) \in \mathcal{N}'$.

Statistics will be computed by summing the weights of experts in subsets of $\mathcal{E}$. If the experts have equal weights, this is equivalent to counting the number of experts. In general, we will sum the weights of experts in a subset $\mathcal{F} \subseteq \mathcal{E}$, and denote the result by $\mu(\mathcal{F})$. Thus $\mu$ is in fact a measure on $\mathcal{E}$, although it is completely determined by the weights of the individual experts $\mu(\{\omega\})$ for $\omega \in \mathcal{E}$. (We are assuming that $\mathcal{E}$ is finite.)

For a given subset $A \subseteq \Lambda$, the characteristic function $\chi_A$ is defined by

$$\chi_A(\lambda) = \begin{cases} 0 & \text{if } \lambda \notin A \\ 1 & \text{if } \lambda \in A. \end{cases}$$

Equality of two functions defined on $\Lambda$ means, of course, that the two functions agree for all $\lambda \in \Lambda$.

Given a space of experts $\mathcal{E}$ and the boolean opinions $X$, we define

$$\bar{m}(A) = \frac{\mu\{\omega \in \mathcal{E} \mid x_\omega = \chi_A\}}{\mu\{\mathcal{E}\}} \quad (5)$$

for every subset $A \subseteq \Lambda$. It is possible to view the values as probabilities on the random variables $\{x(\lambda)\}$. We endow the elements of $\mathcal{E}$ with the prior probabilities $\mu(\{\omega\})/\mu(\mathcal{E})$, and say that the probability of an event involving a combination of the random variables $x(\lambda)$'s over the sample space $\mathcal{E}$ is the probability that the event is true for a particular sample, where the sample is chosen at random from $\mathcal{E}$ with the sampling distribution given by the prior probabilities. This is equivalent to saying

$$\text{Prob}_{\mathcal{E}}(\text{Event}) = \frac{\mu(\{\omega \in \mathcal{E} \mid \text{Event is true for } \omega\})}{\mu\{\mathcal{E}\}}.$$

With this convention, we see that

$$\tilde{m}(A) = \text{Prob}_{\mathcal{E}}(x(\lambda) = \chi_A(\lambda) \text{ for all } \lambda).$$

In fact, all of the priors and joint statistics of the $x(\lambda)$'s are determined by the full collection of $\tilde{m}(A)$ values. For example,

$$\text{Prob}(x(\lambda_0) = 1) = \sum_{\{A \mid \lambda_0 \in A\}} \tilde{m}(A)$$

and

$$\text{Prob}(x(\lambda_0) = 1 \text{ and } x(\lambda_1) = 1) = \sum_{\{A \mid \lambda_0, \lambda_1 \in A\}} \tilde{m}(A).$$

Further, the full set of values $\tilde{m}(A)$ for $A \subseteq \Lambda$ defines an element $\tilde{m} \in \mathcal{M}'$. To see this, it suffices to check that $\sum \tilde{m}(A) = 1$, which amounts to observing that for every $\omega$, $x_\omega = \chi_A$ for some $A \subseteq \Lambda$.

Many of the quantities in the theory of evidence can be interpreted in terms of simple conditional probabilities on the $x(\lambda)$'s. For example, the belief on a set $A$,

$$\text{Bel}(A) = \sum_{B \subseteq A} m(B)$$

is simply the joint probability that $x(\lambda) = 0$ for $\lambda \notin A$ conditioned on the assumption that $x(\lambda) \neq 0$ for some $\lambda \in \Lambda$. In a similar way, plausibility values

$$\text{Pl}(A) = \sum_{B \cap A \neq \emptyset} m(B) = 1 - \text{Bel}(\bar{A})$$

can be interpreted as disjunctive probabilities, and the commonality values

$$Q(A) = \sum_{A \subseteq B} m(B)$$

are a kind of joint probability.

To recapitulate, we have defined a mapping from $P$ values to $X$ values, and then transformations from $X$ to $\tilde{m}$ and $m$ values. The resulting element $m$, which contains statistics on the $X$ variables, is an element in the space of belief states $\mathcal{M}$ of the of the Dempster/Shafer theory of evidence (Section 2).

## 4. Equivalence with the Dempster/Shafer Rule of Combination

At this point, we have four spaces with binary operations, namely $(\mathcal{N}, \otimes)$, $(\mathcal{N}', \odot)$, $(\mathcal{M}', \oplus')$, and $(\mathcal{M}, \oplus)$. We will now show that these four spaces are closely related. It is not hard to show that the binary operation is, in all four cases, commutative and associative, and that each space has an identity element, so that these spaces are abelian monoids. We also have

**Definition 5:** The map $T: \mathcal{N} \to \mathcal{N}'$, with $(\mathcal{E}, \mu, X) = T(\mathcal{E}, \mu, P)$, is given by equation (4), i.e., $x_\omega(\lambda) = 1$ iff $p_\omega(\lambda) > 0$, and $x_\omega(\lambda) = 0$ otherwise. ∎

There is another mapping U, given by

**Definition 6:** The map $U: \mathcal{N}' \to \mathcal{M}'$ with $\tilde{m} = U(\mathcal{E}, \mu, X)$ given by equation (5), i.e., $\tilde{m}(A) = \mu(\{\omega \in \mathcal{E} \mid x_\omega = \chi_A\})/\mu(\{\mathcal{E}\})$. ∎

We claim that T and U preserve the binary operations. More formally, we show that T and U are homomorphisms of monoids. However, proofs are omitted here; we refer the interested reader to the larger report [12].

**Lemma 2:** T is a homomorphism of $\mathcal{N}$ onto $\mathcal{N}'$.

**Lemma 3:** U is a homomorphism of $\mathcal{N}'$ onto $\mathcal{M}'$.

Recall from Section 2 that the map $V: \mathcal{M}' \to \mathcal{M}$ is also a homomorphism. So we can compose the homomorphisms $T: \mathcal{N} \to \mathcal{N}'$ with $U: \mathcal{N}' \to \mathcal{M}'$ with $V: \mathcal{M}' \to \mathcal{M}$ to obtain the following obvious theorem.

**Theorem:** The map $V \circ U \circ T: \mathcal{N} \to \mathcal{M}$ is a homomorphism of monoids mapping onto the space of belief states $(\mathcal{M}, \oplus)$. ∎

This theorem provides the justification for the viewpoint that the theory of evidence space $\mathcal{M}$ represents the space $\mathcal{N}$ via the representation $V \circ U \circ T$.

The significance of this result is that we can regard combinations of elements in the theory of evidence as combinations of elements in the space of opinions of experts. For if $m_1, \cdots, m_k$ are elements in $\mathcal{M}$ that are to be combined under $\oplus$, we can find respective preimages in $\mathcal{N}$ under the map $V \circ U \circ T$, and then combine those elements using the operation $\otimes$ in the space of opinions of experts $\mathcal{N}$. After all combinations in $\mathcal{N}$ are completed, we project back to $\mathcal{M}$ by $V \circ U \circ T$; the result will be the same as if we had combined the elements in $\mathcal{M}$. The advantage to this procedure is that combinations in $\mathcal{N}$ are conceptually simpler: there are no funny normalizations, and we can regard the combination as Bayesian updatings on the product space of experts.

## 5. An Alternative Method for Combining Evidence

With the viewpoint that the theory of evidence is really simply statistics of opinions of experts, we can make certain remarks on the limitations of the theory.

(1) There is no use of probabilities or degrees of confidence. Although the belief values seem to give weighted results, at the base of the theory, experts only say whether a condition is possible or not. In particular, the theory makes no distinction between an expert's opinion that a label is likely or that it is remotely possible.

(2) Pairs of experts combine opinions in a Bayesian fashion with independence assumptions of the sources of evidence. In particular, dependencies in the sources of information are not taken into account.

(3) Combinations take place over the product space of experts. It might be more reasonable to have a single set of experts modifying their opinions as new information comes in, instead of forming the set of all committees of mixed pairs.

Both the second and third limitations come about due to the desire to have a combination formula which factors through to the statistics of the experts and is application-independent. The need for the second limitation, the independence assumption on the sources of evidence, is well-known (see, e.g., [13]). Without incorporating much more complicated models of judgements under multiple sources of knowledge, we can hardly expect anything better.

The first objection, however, suggests an alternate formulation which makes use of the probabilistic assessments of the experts. Basically, the idea is to keep track of the density distributions of the opinions in probability space. Of course, complete representation of the distribution would amount to recording the full set of opinions $p_\omega$ for all $\omega$. Instead, it is more reasonable to approximate the distribution by some parameterization, and update the distribution parameters by combination formulas.

We present a formulation based on normal distributions of logarithms of updating coefficients. Other formulations are possible. In marked contrast to the Dempster/Shafer formulation, we assume that all opinions of all experts are nonzero for every label. That is, instead of converting opinions into boolean statements by test for zero, we will assume that all the values are nonzero, and model the distribution of their strengths.

In a manner similar to [14], set

$$L(\lambda | s_i) = \log\left[\frac{\text{Prob}(\lambda | s_i)}{\text{Prob}(\lambda)}\right],$$

where $\text{Prob}(\lambda | s_i)$ is a probability of label $\lambda$ being the correct labeling, among labeling situations, conditioned on some information $s_i$ shared by the collection of experts $\mathcal{E}_i$. (Note, incidentally, that the $L(\lambda | s_i)$ values are not the so-called "log-likelihood ratios"; in particular, they can be both positive and negative). Using some fairly standard assumptions in Bayesian updating, (see [14]), we obtain

$$\log[\text{Prob}(\lambda | s_1, \cdots, s_k)] \approx$$

$$c + \log[\text{Prob}(\lambda)] + \sum_{i=1}^{k} L(\lambda | s_i),$$

where $c$ is a constant independent of $\lambda$ (but not of $s_1, \cdots, s_k$).

The consequence of this formula is that if certain independence assumptions hold, and if $\text{Prob}(\lambda)$ and $L(\lambda | s_i)$ are known for all $\lambda$ and $i$, then the approximate values $\text{Prob}(\lambda | s_1, \cdots, s_k)$ can be determined.

Accordingly, we introduce a space which we term "logarithmic opinions of experts." For convenience, we will assume that experts have equal weights. An element in this space will consist of a set of experts $\mathcal{E}_i$, and a collection of opinions $Y_i = \{y_\omega^{(i)}\}_{\omega \in \mathcal{E}_i}$. Each $y_\omega^{(i)}$ is a map, and the component $y_\omega^{(i)}(\lambda)$ represents expert $\omega$'s estimate of $L(\lambda | s_i)$:

$$y_\omega^{(i)} : \Lambda \to \mathbb{R}, \quad y_\omega^{(i)}(\lambda) \approx L(\lambda | s_i).$$

Note that the experts in $\mathcal{E}_i$ all have knowledge of the information $s_i$, and that the estimated logarithmic coefficients $L(\lambda | s_i)$ can be positive or negative. In fact, since the experts do not necessarily have precise knowledge of the value of $\text{Prob}(\lambda)$, but instead provide estimates of log's of ratios, the estimates can lie in an unbounded range.

Combination in the space of logarithmic opinions of experts is defined much the same as our earlier combination formulas, except that now consensus opinions are derived by adding component opinions. Specifically, the combination of $(\mathcal{E}_1, Y_1)$ and $(\mathcal{E}_2, Y_2)$ is $(\mathcal{E}_1 \times \mathcal{E}_2, Y)$, where

$$Y = \{y_{(\omega_1, \omega_2)}\}_{(\omega_1, \omega_2) \in \mathcal{E}},$$

and

$$y_{(\omega_1,\omega_2)}(\lambda) = y^{(1)}_{\omega_1}(\lambda) + y^{(2)}_{\omega_2}(\lambda).$$

Next, in analogy with our map to a statistical space (Section 3.2), we can define a space which might be termed the "parameterized statistics of logarithmic opinions of experts." Elements in this space will consist of pairs $(\bar{u},C)$, where $\bar{u}$ will be a mean vector in $\mathbb{R}^n$ and $C$ is a symmetric $n$ by $n$ covariance matrix. To project from the space of logarithmic opinions to the space of parameterized statistics, define $u_i$ to be the average value of $y_\omega(\lambda_i)$ over $\omega \in \mathcal{E}$, where $\Lambda = \{\lambda_1, \cdots, \lambda_n\}$ is a fixed ordering of the elements in the label set. Then the vector $\bar{u}$ is defined by $\bar{u} = (u_1, \cdots, u_n)$. Likewise, define $c_{ij}$ as the average value of $(y_\omega(\lambda_i) - u_i) \cdot (y_\omega(\lambda_j) - u_j)$ over $\omega \in \mathcal{E}$, and set C equal to the matrix whose $i,j$-th component is given by $c_{ij}$.

Combinations in the space of statistics must be defined in such a way that the map from the collections of opinions to the mean and covariances forms a homomorphism. We are led, after some calculation, to the definition:

$$(\bar{u}^{(1)},C^{(1)}) \oplus (\bar{u}^{(2)},C^{(2)}) = (\bar{u}^{(1)}+\bar{u}^{(2)}, C^{(1)}+C^{(2)}).$$

That is, since the components are added on the product space of experts, the means and covariances separately add. An extension to the case where $\mathcal{E}_1$ and $\mathcal{E}_2$ have nonequal total weights is straight-forward.

To interpret a state $(\bar{u},C)$ in the space of parameterized statistics, we must remember the origin of the logarithmic-opinion values. Specifically, after $k$ updating iterations combining information $s_1$ through $s_k$, the updated vector $\bar{y} = (y_1, \cdots, y_n) \in \mathbb{R}^n$ is an estimate of the sum of the logarithmic coefficients,

$$y_j \approx \sum_{i=1}^{k} L(\lambda_j | s_j).$$

The a posteriori probability of a label $\lambda_j$ is high if the corresponding coefficient $y_j + \log[\text{Prob}(\lambda_j)]$ is large in comparison to the other components $y_j + \log[\text{Prob}(\lambda_j)]$.

Since the state $(\bar{u},C)$ represents a multinormal distribution in the log-updating space, we can transform this distribution to a density function for a posteriori probabilities. The vector $\bar{u}$ represent the center of the distribution (before bias by the priors). The spread of the distribution is given by the covariance matrix, which can be thought of as defining an ellipsoid in $\mathbb{R}^n$ centered at $\bar{u}$. The exact equation of the ellipse can be written implicitly as:

$$(\bar{y}-\bar{u})^T C^{-1} (\bar{y}-\bar{u}) = 1.$$

This ellipse describes a "one sigma" variation in the distribution, representing a region of uncertainty of the logarithmic opinions; the distribution to two standard deviations lies in a similar but enlarged ellipse. The eigenvalues of $C$ give the squared lengths of the semi-major axes of the ellipse, and are accordingly proportional to degrees of confidence in the corresponding directions. Bias by the prior probabilities simply adds a fixed vector, with components $\log[\text{Prob}(\lambda_j)]$, to the ellipse, thereby translating the distribution. We seek an axis $j$ such that the components $y_j$ of the vectors $y$ lying in the translated ellipse are relatively much larger than other components of vectors in the ellipse. In this case, the preponderant evidence is for label $\lambda_j$.

Clearly, the combination formula is extremely simple. Its greatest advantage over the Dempster/Shafer theory of evidence is that only $O(n^2)$ values are required to describe a state, as opposed to the $2^n$ values used for a mass distribution in $M$. The simplicity and reduction in numbers of parameters has been purchased at the expense of an assumption about the kinds of distributions that can be expected. However, the same assumption allows us to track probabilistic opinions (or actually, the logarithms), instead of converting all opinions into boolean statements about possibilities.

## 6. Conclusions

We have shown how the theory of evidence may be viewed as a representation of a space of opinions of experts, where opinions are combined in a Bayesian fashion over the product space of experts. By "representation", we mean something very specific — namely, that there is a homomorphism mapping from the space of opinions of experts onto the Dempster/Shafer theory of evidence space. This map fails to be an isomorphism (which would imply equivalence of the spaces) only insofar as it is many-to-one. In this way the state in the theory of evidence represents a corresponding collection of elements.

Furthermore, combination in the space of opinions of experts, as defined in Section 3, leads to combination in the theory of evidence space. This allows us to implement combination in a somewhat simpler manner, since the formulas for combination without the normalization are simpler than the more standard formulas, and also permits us to view combination in the theory of evidence space as the tracking of statistics of opinions of experts as they combine information in a pairwise Bayesian fashion over the product space of experts.

From this viewpoint, we can see how the Dempster/Shafer theory of evidence accomplishes its goals, and what independence assumptions are needed. Degrees of support for a proposition, belief, and plausibilities, are all measured in terms of joints and disjunctive probabilities over a set of experts who are naming possible labels given current information. The problem of ambiguous knowledge versus uncertain knowledge, which is frequently described in terms of "withholding belief," can be viewed as two different distributions of opinions. In particular, ambiguous knowledge can be seen as observing high densities of opinions on particular disjoint subsets, whereas uncertain knowledge corresponds to unanimity of opinions, where the agreed upon opinion gives many possibilities. Finally, instead of performing Bayesian updating, a *set* of values are updated in a Bayesian fashion over the product space, which results in non-Bayesian formulas over the space of labels.

In meeting each of these goals, the theory of evidence invokes compromises that we might wish to change. For example, in order to track statistics, it is necessary to model the distribution of opinions. If these opinions are probabilistic assignments over the set of labels, then the distribution function will be too complicated to retain precisely. The Dempster/Shafer theory of evidence solves this problem by simplifying the opinions to boolean decisions, so that each expert's opinion lies in a space having $2^n$ elements. In this way, the full set of statistics can be specified using $2^n$ values. We have suggested an alternate method, which retains the probability values in the opinions without converting them into boolean decisions, and requires only $O(n^2)$ values to model the distribution, but fails to retain full information about the distribution. Instead, our method attempts to approximate the distribution of opinions with a Gaussian function.

## Acknowledgements

This research was supported by Office of Naval Research Grant N00014-85-K-0077 and NSF Grant DCR-8403300. This is a conference report version of a longer paper, which appears as NYU Robotics Research Report Number 57 and can be obtained from the author at 251 Mercer Street, New York, New York, 10012, or by a request sent to "hummel@nyu.arpa". Many useful suggestions were given by Tod Levitt. We appreciate also the helpful comments given by George Reynolds, Deborah Strahman, and Jean-Claude Falmagne.

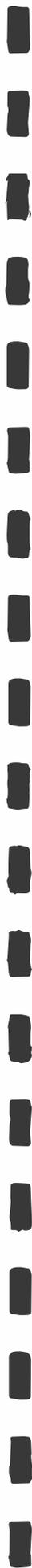